\def\year{2017}\relax
\documentclass[letterpaper]{article}
\usepackage{aaai17}
\usepackage{times}
\usepackage{helvet}
\usepackage{courier}

\usepackage{graphicx}
\usepackage{float}
\usepackage{algorithm}
\usepackage[noend]{algpseudocode}
\usepackage{etoolbox}
\usepackage{subcaption}
\usepackage{tikz}
\usepackage{amsmath}
\usepackage{natbib}

\begin{document}
%
\title{Parallel Chromatic MCMC with Spatial Partitioning}
\author{
  Jun Song \and David A. Moore \\
  Computer Science Division\\
  University of California at Berkeley\\
  Berkeley, CA 94720 \\
  juns123@berkeley.edu, dmoore@cs.berkeley.edu \\
  }
\maketitle
\begin{abstract}
We introduce a novel approach for parallelizing MCMC inference in
models with spatially determined conditional independence
relationships, for which existing techniques exploiting graphical
model structure are not applicable. Our approach is motivated by
a model of seismic events and signals, where events detected in
distant regions are approximately independent given those in intermediate
regions. We perform parallel inference by coloring a factor graph
defined over regions of latent space, rather than individual model variables.
Evaluating on a model of seismic event detection, we achieve
significant speedups over serial MCMC with no degradation in inference quality.
\end{abstract}

\section{Introduction} 
Probabilistic modeling is becoming an increasingly important framework for designing machine learning algorithms. However, scaling inference to large datasets is still a major challenge. Methods such as Markov chain Monte Carlo (MCMC) can require many steps to mix, especially for complex models on large datasets of practical interest. A natural solution is to exploit additional computational resources via parallel inference, but existing methods for parallel MCMC often depend on graphical model structure that may not be present in all problems. 

This paper proposes a new approach for parallel  inference in models with conditional independence relationships induced by spatial structure rather than a fixed graphical model. Our approach is inspired by an application to seismic event detection \citep{NIPS2010_4100}, and we use a toy model of seismic events and signals as the running example to illustrate the method.
We partition a continuous latent space into regions, inducing a conditional independence structure on the {\em sets} of variables taking values in each region, rather than individual variables themselves. Motivated by Chromatic Gibbs Sampling \citep{Gonzalez+al:aistatspgibbs}, we color the regions using an induced factor graph, and run inference in parallel within all regions of a given color. The spatial partitioning naturally allows for data subsampling, since the Markov blanket \citep{Pearl:1988:PRI:52121} of each region involves only a local subset of observed data. We  further improve mixing by applying dynamic partitioning to avoid boundary effects between spatial regions. Although this paper focuses on an application to seismic event detection, the technique is applicable more generally to any model in which objects or latent structures are inferred from "local" surroundings in a large space. This potentially includes, among others, large-scale visual recognition, object detection from aerial or satellite imagery, radar tracking, modeling long time series such as speech or video, and simultaneous localization and mapping (SLAM).

We briefly review MCMC inference and graph coloring-based parallelization in the next section, before introducing the toy model of seismic events and illustrating our approach for parallel inference by spatial partitioning. Evaluating our approach with a comparison to serial MCMC, and to a na\"ive partitioning that does not preserve the correct stationary distribution, we observe significant speedups from parallelism without compromising the quality of the resulting inferences, as measured by precision, recall, and mean error in the location of recovered events. 

\section{Background}

MCMC techniques are often applied to solve integration and optimisation problems in large dimensional spaces, including inference in probabilistic models. Most practically applied MCMC chains are constructed using the framework of the Metropolis-Hastings (MH) algorithm. An MH step of invariant distribution $\pi(x)$ and proposal distribution $q(x^*|x)$ involves sampling a candidate value $x^*$ given the current value $x$ according to $q(x^*|x)$. The Markov
chain then moves towards $x^*$ with acceptance probability $A(x, x^*) = \min \{ 1, \frac{\pi(x^*)q(x|x^*)}{\pi(x)q(x^*|x)} \}$, and otherwise remains at $x$ \citep{Jordan2003b}. If the proposal distribution is symmetric, $q(x^*| x) = q(x|x^*)$, the acceptance probability reduces to $\alpha(x^*|x) = \min \{ 1, \frac{\pi(x^*)}{\pi(x)}\}$. 

Gibbs sampling is the special case of Metropolis Hasting where the proposal distributions are the posterior conditionals. The acceptance probability of a Gibbs move is always $1$ and thus all proposals are accepted.

Many methods have been proposed to parallelize MCMC inference. We build on the work of  \citet{Gonzalez+al:aistatspgibbs}, which considers a Markov random field (MRF) with a $k$-coloring such that each vertex is assigned one of k colors and adjacent vertices have different colors. Let $k_i$ denote the variables in color $i$. Then the Chromatic sampler simultaneously draws new values for all variables in $k_{i}$ before proceeding to $k_{i+1}$. The $k$-coloring of the graph ensures that all variables within a color are conditionally independent given the variables in the remaining colors and can therefore be sampled independently in parallel. 

\begin{algorithm}
\begin{algorithmic}[1]
\caption{Chromatic Gibbs Sampler}\label{euclid}
\State Input: k-color MRF
\For {each of $k$ colors $k_i$: $i \in 1 \dots k$}
\For {\textbf{all} $X_i \in k_i$ in the $i^{th}$ color} \textbf{in parallel}
\State $\textit{Execute Gibbs Update}$
\State $\hspace{0.3cm} X_j^{(t+1)} \sim \pi(X_j | X_{N_j \in k < i}^{(t+1)}, X_{N_j \in k > i}^{(t)})$
\EndFor
\EndFor
\end{algorithmic}
\end{algorithm}

However, the Chromatic Gibbs sampler depends on a fixed factor graph structure and cannot exploit additional independences implied by spatial relationships between variables.

\section{Generative Model of Seismic Signals} 
We motivate our proposed inference approach in a toy model of seismic event detection, which we use as a running example. Our model describes an unknown number of seismic events in a world with one spatial dimension, so that each event $i$ is described by a tuple $e_i = (x_i, t_i)$ giving its scalar location and time, as well as additional latent variables including the arrival times of event signals at each of four stations. The signal observed at each station includes contributions from events as well as a background noise process. The inference problem is to recover the number of events, and their space-time positions, given noisy observed signals.

\textbf{Event prior} The number of events are generated from a Poisson prior 
$P(|e|) = \frac{(\lambda \cdot T)^{|e|} \exp (-\lambda \cdot T)}{|e|!}$, where $\lambda$ is the event generation rate, $T$ is the time span under consideration, $e$ is the set of events, $|e|$ is the size of the set (number of events). The location of the event is uniformly distributed up to a maximum $x_{max}$. The time of the event is uniformly distributed between $0$ and $T$.

\textbf{Event Detections}
The arrival time $a_{ij}$ of event $i$ at station $j$ is assigned by a Gaussian distribution with mean equal to 
$t_i + \frac{|x_i - x_{s_j}|}{v}$ where $x_{s_j}$ is the station location and $v$ is the velocity of the seismic wave ($v=2$ in our experiments): 

$P(a_{ij}|x_i, t_i) = \mathcal{N} \left(t_i + \frac{|x_i - x_{s_j}|}{v}, \sigma ^2\right)$

The max travel time $\tau_{max} = x_{max}/v$ is defined as the time a seismic wave takes to travel from one end of the space to the other.

The signal generated by event $i$ at station $s_j$ lasts for a fixed duration $t_s$, covering the period $[a_{ij}, a_{ij} + t_s]$, and is sampled to be iid Gaussian, with larger variance than the background noise process. 

\textbf{Observed Signals} The signal observed at each station is the sum of signals from arriving events, added together at the appropriate time offsets along with iid Gaussian background noise. Since each event signal is Gaussian, as is the background noise, the observed signal is itself Gaussian conditioned on the arrival times $a$. 

\begin{figure}[H]
  \centering
  \includegraphics[height=4.0cm, width=2.7cm]{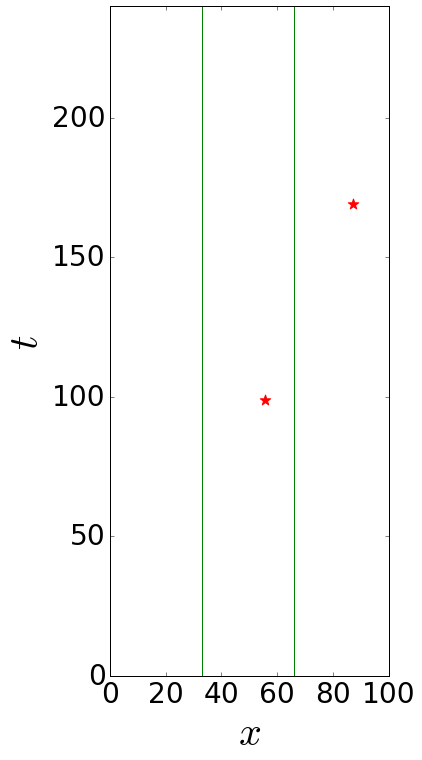}
  \caption{Example of two events located in space-time. Event $e_1$ can be described by $t_1=169$, $x_1 = 87$. Events $e_2$ can be described by $t_2=99$, $x_2=56$. Locations of detecting stations are shown by vertical lines, $x=0$, $33$, $66$, and $100$.}
  \label{fig:two_events}
\end{figure}

\begin{figure}[H]
\centering
\captionsetup[subfigure]{labelformat=empty}
\begin{subfigure}{.23\textwidth}
  \centering
  \includegraphics[width=1.0\linewidth]{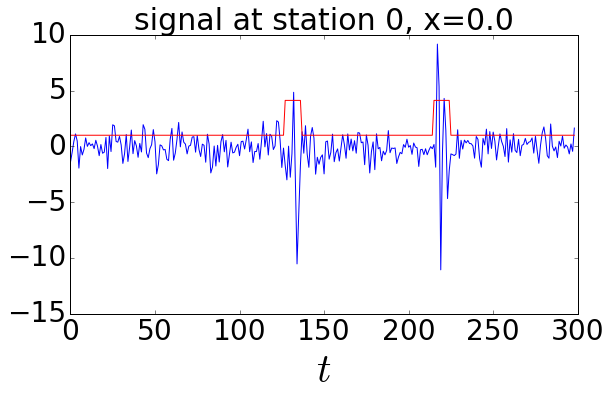}
\end{subfigure}%
\begin{subfigure}{.23\textwidth}
  \centering
  \includegraphics[width=1.0\linewidth]{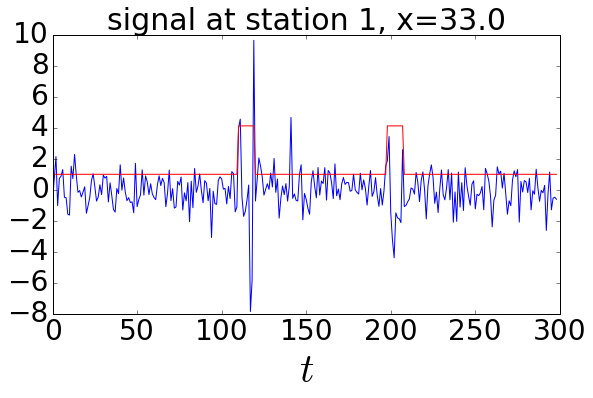}
\end{subfigure}
\begin{subfigure}{.23\textwidth}
  \centering
  \includegraphics[width=1.0\linewidth]{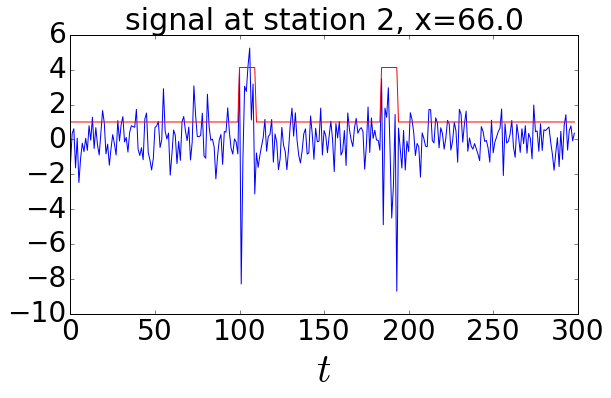}
\end{subfigure}%
\begin{subfigure}{.23\textwidth}
  \centering
  \includegraphics[width=1.0\linewidth]{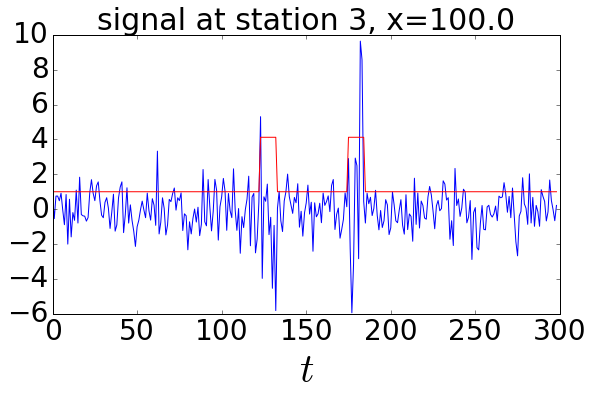}
\end{subfigure}
\caption{Signals (blue) generated by events in Figure ~\ref{fig:two_events}, with arrival times dependent on the event--station distance. Red indicates the marginal standard deviation.}
\label{fig:signals}
\end{figure}
The overall probability of any set of events $e$ and signals $s$ can be written as

\begin{align*}
    P(e, a, s) &= P(e) \left(\prod_j^{|s|} \left(\prod_i^{|e|} P(a_{ij}|e_i) \right) P(s_j|a, e)\right)\\
    P(e) &= P(|e|)\prod_{i=1}^{|e|} P(x_i)P(t_i)
\end{align*}

where the signal model $P(s_j|e, a)$ is a zero-mean, diagonal Gaussian density with variance at each timestep given by the sum of variances from the noise process and any arriving signals. Figures \ref{fig:two_events} and \ref{fig:signals} show a sample from this model, including space-time locations of two events and the signals they generate. 

We use Metropolis Hastings to infer the posterior on events given observed signals. A serial inference algorithm for this model involves several types of MH proposals:

\textbf{Birth and Death Moves:} We propose birthing a new event from a uniform distribution, or killing an existing event.

\textbf{Event Location Move:} We propose moving a random event by a Gaussian offset in space and time.

\textbf{Arrival Time Move: } We propose changing the arrival time of a random event at a random station by a Gaussian offset.

\textbf{Joint Events Move: } The locations of nearby events are coupled by the observed signals, potentially including multiple joint modes from aliasing effects. We analyze the station geometry to jointly propose moving a pair of random events to another high probability mode (Figure \ref{overflow}).

\section{Parallel Metropolis-Hastings Inference}

\makeatletter
\let\OldStatex\Statex
\renewcommand{\Statex}[1][3]{%
  \setlength\@tempdima{\algorithmicindent}%
  \OldStatex\hskip\dimexpr#1\@tempdima\relax}
\makeatother

\begin{algorithm}[H]
\begin{algorithmic}[1]
\caption{Serial Metropolis Hastings}\label{euclid}
\State Input: world x
\For {$j$ in range($k$) steps}
\State $ \textit{pick a type of move, make proposal} \hspace{0.2cm} x^{cand}$
\State $\textit{accept move with probability: } $
\Statex $\alpha(x^{cand}|x^{(j-1)}) =$
\Statex $\min \{1,\frac{q(x^{(j-1)}|x^{cand})\pi(x^{cand})}{q(x^{cand}|x^{(j-1)})\pi({x^{(j-1)})}} \}$
\If {accepted}
\State $x^{(j)} = x^{cand}$
\Else
\State $x^{(j)} = x^{(j-1)}$
\EndIf
\EndFor
\end{algorithmic}
\end{algorithm}

Running inference on real seismic event signals serially can take a long time. In a more complicated model with real event signals from hundreds of stations all over the space, the inference process on a $24$-hour period signals from roughly $100$ stations will take weeks or even a month.

However, this model contains conditional independences that we can exploit for parallel inference. To a first approximation, the set of events occurring during a given time period $[t_1, t_2]$ can be inferred using only the signals from that time period. In fact it is also necessary to consider signals up to the maximum travel time $\tau_{max}$ following the time period in question, since an event at time $t$ may not be detected until $t+\tau_{max}$. Since these additional signals may also include arrivals from events in the following period $[t, t+\tau_{max}]$, the posterior on events is coupled by the observed signals and does not decompose cleanly over adjacent time periods. However, the events in a given time period are independent of others {\em conditioned} on those occurring within the maximum travel time, i.e., conditioned on those events whose signals could directly ``explain away'' arrivals from the time period of interest.

Note that this independence is induced by the locations $(x_i, t_i)$ of the inferred events, and is not present {\em a priori} in the factorization of the our model. Concretely, each timestep of signal depends {\em a priori} on {\em every} inferred event $e_i$, and in fact the exchangeability of events in our model introduces a labeling symmetry, so that the event variables $e_i$ are strongly coupled by the observed signals. Therefore methods such as Chromatic Gibbs \citep{Gonzalez+al:aistatspgibbs} that exploit graphical model structure are not directly applicable.

 Our contribution is to apply chromatic methods to an {\em induced} factor graph constructed from a spatial partition. We construct a factor graph by partitioning the continuous space and design  parallel Metropolis Hastings algorithms that color these partitions and run inference on regions of the same color in parallel. 

\subsection{Partitioning the World}

We divide the latent event space into regions corresponding to fixed time periods $r^0, r^1, r^2 \dots$. The region $r^n$ describes a two dimensional space, with $x$ ranging from $0$ to $x_{max}$, $t$ ranging from $nl$ to $(n + 1)l$ where $l$ is the length of the region. We will also overload notation to let $r^n$ refer to the set of events occurring in that region. 

We considered partitioning by both space and time, instead of time alone; however, for this simple model there is no clear gain because the conditional independence is induced by the maximum travel time. A spatial partition would be natural if our model included attenuation, so that events are only detected by stations up to some maximum distance. In general it might be advantageous to consider more complex partitions across multiple latent dimensions, depending on the conditional independence structure of the model. 

We also divide observed signals into time periods, written as $s^1, s^2 \dots$. The set of signals $s^n$ describes signals collected by stations from $t = nl$ to $(n + 1)l$.

When $l \ge \tau_{max}$, the Bayesian network of signals, regions is shown in the Figure~\ref{fig:bi_color}.

Using the property of Markov blanket \citep{Pearl:1988:PRI:52121}, $r^{n+2}$ are independent from $r^n$ when conditioned on the Markov blanket of $r^n$, which contains $r^{n+1}$ and $s^n, s^{n+1}$. Thus, we can bi-color the graph by assigning every two regions to the same color. With this coloring, regions of one color are mutually independent when conditioning on regions of other colors and the set of signals.

\tikzstyle{blue} = [draw, fill=blue!20, circle, node distance=1cm]
\tikzstyle{red} = [draw, fill=red!20, circle, node distance=1cm]
\tikzstyle{grey} = [draw, fill=black!10, circle, node distance=1cm]

\begin{figure}[H]
\centering
\begin{tikzpicture}

\draw (0,0) node[blue, minimum size = 0.5cm] (R_0) {$r^0$};
\draw (1.5,0) node[red, minimum size = 0.5cm] (R_1){$r^1$};
\draw (3,0) node[blue, minimum size = 0.5cm] (R_2){$r^2$};
\draw (4.5,0) node[red, minimum size = 0.5cm] (R_3){$r^3$};

\draw (0,-1.5) node[grey, minimum size = 0.5cm] (S_0){$s^0$};
\draw (1.5,-1.5) node[grey, minimum size = 0.5cm] (S_1) {$s^1$};
\draw (3,-1.5) node[grey, minimum size = 0.5cm] (S_2) {$s^2$};
\draw (4.5,-1.5) node[grey, minimum size = 0.5cm] (S_3) {$s^3$};
\draw (6,-1.5) node[grey, minimum size = 0.5cm] (S_4) {$\dots$};

\draw[->]  (R_0) -- (S_0);
\draw[->]  (R_0) -- (S_1);
\draw[->]  (R_1) -- (S_1);
\draw[->]  (R_1) -- (S_2);
\draw[->]  (R_2) -- (S_2);
\draw[->]  (R_2) -- (S_3);
\draw[->]  (R_3) -- (S_3);
\draw[->]  (R_3) -- (S_4);

\end{tikzpicture}
\caption{Factor graph when $l \ge \tau_{max}$, Bi-color}
\label{fig:bi_color}
\end{figure}
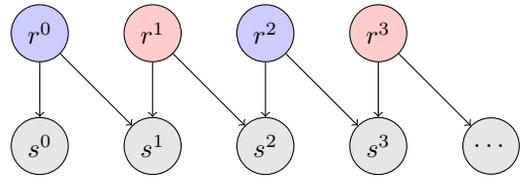

We could also decrease the region size to $ \frac{1}{2}\tau_{max} \le l < \tau_{max} $ and three-color the regions, but a two-coloring will always allow the most parallelism.

\subsection{Naive Parallel Metropolis Hastings}

\begin{figure}[H]
\centering
\begin{tikzpicture}

\draw (0,0) node[blue, minimum size = 0.5cm] (R_0) {$r^0$};
\draw (3,0) node[blue, minimum size = 0.5cm] (R_1){$r^1$};
\draw (6,0) node[blue, minimum size = 0.5cm] (R_2){$r^2$};

\draw (0,-1.5) node[grey, minimum size = 0.5cm] (S_0){$s^0$};
\draw (1.5,-1.5) node[grey, minimum size = 0.5cm] (S_1) {$s^1$};
\draw (3,-1.5) node[grey, minimum size = 0.5cm] (S_11) {$s^1$};
\draw (4.5,-1.5) node[grey, minimum size = 0.5cm] (S_2) {$s^2$};
\draw (6,-1.5) node[grey, minimum size = 0.5cm] (S_22) {$s^2$};
\draw (7.5,-1.5) node[grey, minimum size = 0.5cm] (S_3) {$\dots$};

\draw (2.25,-1.5) ellipse (1.3cm and 0.8cm);
\draw (5.25,-1.5) ellipse (1.3cm and 0.8cm);

\draw[->]  (R_0) -- (S_0);
\draw[->]  (R_0) -- (S_1);
\draw[->]  (R_1) -- (S_11);
\draw[->]  (R_1) -- (S_2);
\draw[->]  (R_2) -- (S_22);
\draw[->]  (R_2) -- (S_3);

\end{tikzpicture}
\caption{Factor graph of Naive Parallel Metropolis-Hastings}
\label{fig:naive_factor_graph}
\end{figure}
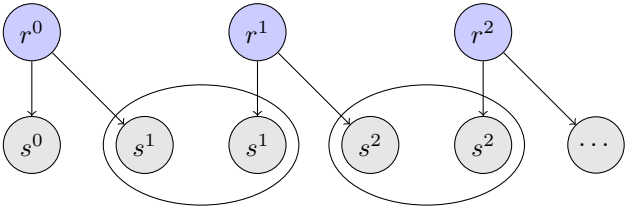

As a simple baseline, we consider the Naive Parallel algorithm, which simply performs inference on all time periods in parallel, so that each block of signals $s^n$ is explained by events $r^n$ and also, independently, by events $r^{n-1}$ occurring during the previous time period. As seen in Figure~\ref{fig:naive_factor_graph}, this inference method breaks the factor graph and does not converge to the correct overall stationary distribution.

\begin{algorithm}[H]
\begin{algorithmic}[1]
\caption{Naive Parallel Metropolis Hastings}\label{euclid}
\State Input: static world that are divided into regions of length $\tau_{max}$, each region is written as $x_m$
\For {$\textbf{all} $ regions $m$ } \textbf{in parallel}
\For {$j$ in range($k$) steps}
\State $\textit{pick a move randomly, make proposal} \hspace{0.3cm} x_m^{cand}$
\State $\textit{accept move with probability:}$ 
\Statex $\alpha(x_m^{cand}|x_m^{(j-1)})= $
\Statex $\min \{ 1,\frac{q(x_m^{(j-1)}|x_m^{cand})\pi(x_m^{cand})}{q(x_m^{cand}|x_m^{(j-1)})\pi(x_m^{(j-1)})} \}$
\If {accepted}
\State  $x_m^{(j)} = x_m^{cand}$
\Else
\State $x_m^{(j)} = x_m^{(j-1)}$
\EndIf

\EndFor
\EndFor
\end{algorithmic}
\end{algorithm}

\begin{figure}
\raggedright
\captionsetup[subfigure]{labelformat=empty}
\begin{subfigure}{.15\textwidth}
  \centering
  \includegraphics[width=2.8cm,height=4.5cm]{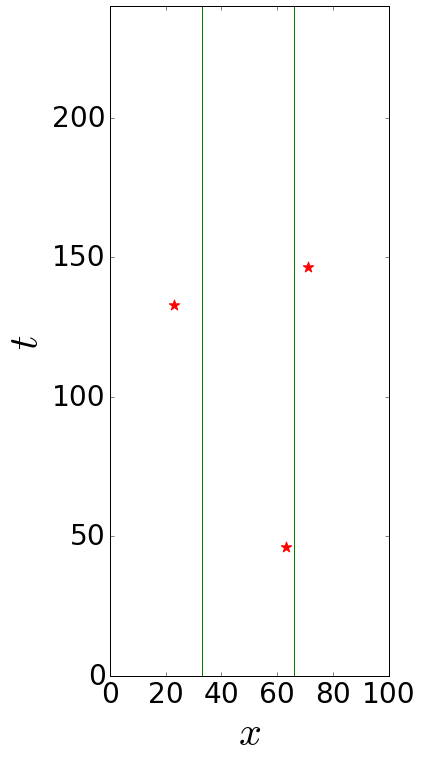}
  \caption{}
\end{subfigure}%
\begin{subfigure}{.15\textwidth}
  \centering
  \includegraphics[width=2.8cm,height=4.5cm]{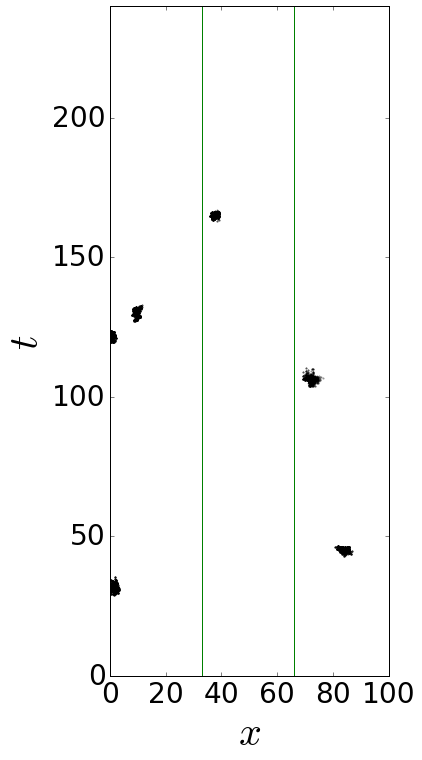}
  \caption{}
\end{subfigure}
\captionsetup[subfigure]{labelformat=empty}
\begin{subfigure}{.15\textwidth}
  \centering
  \includegraphics[width=2.8cm,height=4.5cm]{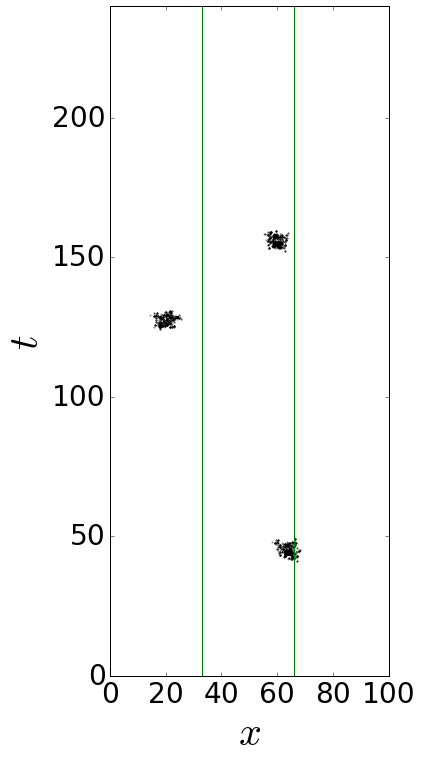}
  \caption{}
\end{subfigure}%
\caption{Red stars in the left sub-figure are actual events. The middle sub-figure shows $1000$ samples of Naive Parallel Metropolis-Hastings. The right sub-figure shows $1000$ samples of Chromatic Dynamic Metropolis-Hastings. }
\label{fig:naive_results}
\end{figure}

For a world that has $n$ regions, this approach can achieve up to $n$ times speed up. However, because it imposes additional independence assumptions, it is likely to generate multiple events to explain a true event, especially when the true event is near the region boundary (See Figure \ref{fig:naive_results}). To fix this problem, we design a Chromatic Metropolis Hastings algorithm.

\subsection{Chromatic Metropolis Hastings}
\begin{figure}[H]
\centering
\captionsetup[subfigure]{labelformat=empty}
\begin{subfigure}{.23\textwidth}
  \centering
  \includegraphics[width=2.8cm, height=4.0cm]{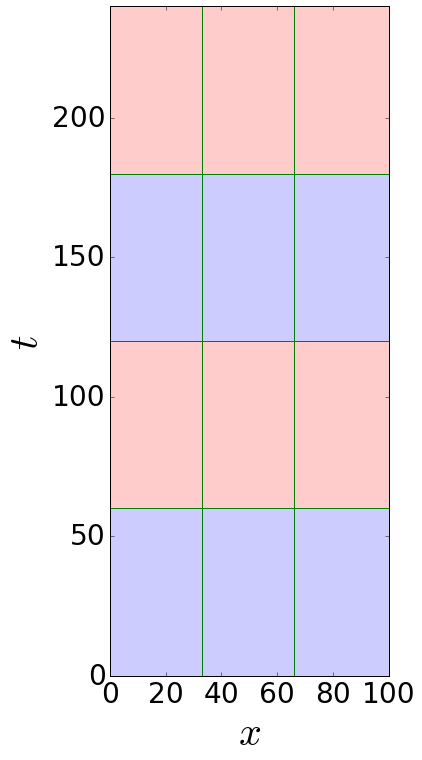}
  \caption{Actual Coloring $l = \tau_{max}$}
\end{subfigure}%
\begin{subfigure}{.25\textwidth}
  \centering
  \includegraphics[width=4.2cm, height=4.5cm]{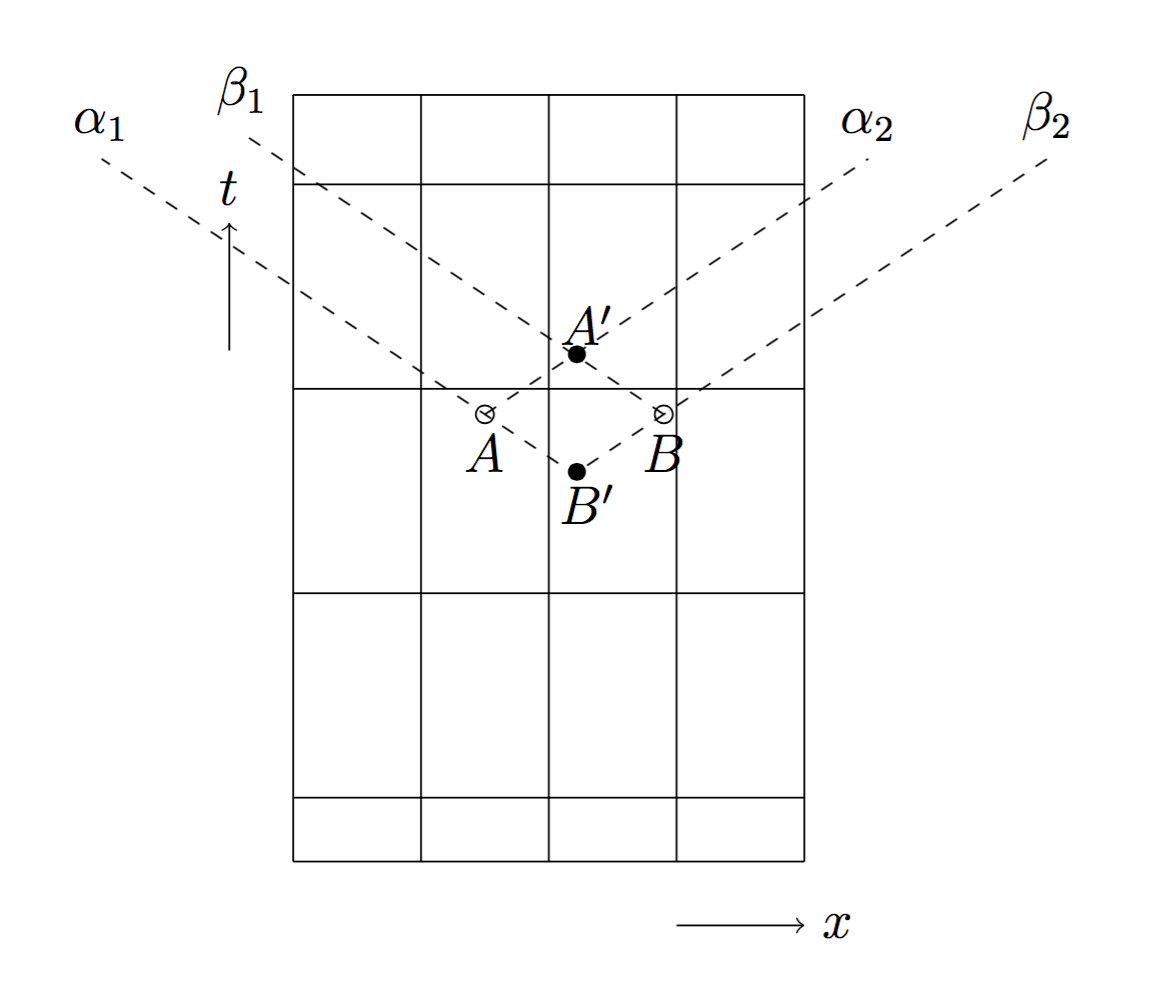}
  \caption{ Local Maxima and Another \\ High Probability Mode}
\end{subfigure}
\caption{}
\label{fig:actual_coloring}
\end{figure}

In our implementation, we assign $l$ (region length) to $\tau_{max}$ and bi-color the world. The factor graph of this setting is in Figure~\ref{fig:bi_color}. The actual coloring of this setting is in Figure~\ref{fig:actual_coloring}.

\subsubsection{Chromatic Metropolis Hastings with Static Coloring}
The Chromatic Metropolis Hastings algorithm we designed is based on a static two colored world, which means the boundaries of each colored region and the order of region colors stay the same throughout the inference process. The method to color the world is as described above. Inference alternates between the two colors, running a sequence of MH updates on all red regions in parallel, then on all blue regions given the red regions, and so on. For a world with $n$ regions, Chromatic Metropolis Hastings can get $\frac{n}{2}$ times speed up ideally.

\begin{algorithm}[H]
\begin{algorithmic}[1]
\caption{Chromatic Metropolis Hastings with Static Coloring}\label{euclid}
\State Input: a two-colored world, initialize $X^{(0)}$ to be empty, $X^{(i)}$ will store region events after the completion of $i^{th}$ full round inference on blue or red regions. $x^{(j)}$ stores temporary region events during a single round of inference.
\While {$i <$ max iterations}
\For {$A$ in $\{ $blue, red$ \}$}
\For {\textbf{all} regions $m$ in color $A$} \textbf{in parallel}
\State $x_m^{(0)} = X_m^{(i-1)}$
\For {$j$ in range($k$)}
\State \textit{pick a move, make proposal} $x_m^{cand}$
\State $\textit{accept move with probability: } $
\Statex $\alpha (x_m^{cand}, X_{\neg m}^{(-1)} | x_m^{(j-1)}, X_{\neg m}^{(-1)}) = $ 

$\min \{1, \frac{q(x_m^{(j-1)}, X_{\neg m}^{(-1)} | x_m^{cand}, X_{\neg m}^{(-1)})\pi(x_m^{cand}, X_{\neg m}^{(-1)})}{q(x_m^{cand}, X_{\neg m}^{(-1)} | x_m^{(j-1)}, X_{\neg m}^{(-1)})\pi(x_m^{(j-1)}, X_{\neg m}^{(-1)}) }\}$
\If {accepted}
\State $x_m^{(j)} = x_m^{cand}$
\Else 
\State $x_m^{(j)} = x_m^{(j-1)}$
\EndIf
\EndFor
 \State $X_m^{(i)} = x_m^{(k)}$
\EndFor
\EndFor
\State $i = i+1$
\EndWhile
\end{algorithmic}
\end{algorithm}

When doing inference using Chromatic Metropolis Hastings, all regions that run in parallel are independent from one another, conditioned on the contents of the remaining regions. Because birth and death moves allow for events to be removed from one region and reborn in another, in principle, the joint chain is ergodic and has the correct stationary distribution. However, in practice the inability to move events across region boundaries leads to difficulties in mixing, especially for events whose posteriors straddle a boundary. And we are likely to wrongly infer near-boundary events to be in their neighbor regions. In addition, we cannot apply joint moves to pairs of nearby events separated by region boundaries, so they may get stuck in suboptimal modes. To fix these problems, we slightly modify the above algorithm and come up with the algorithm in the following section.

\subsubsection{Chromatic Metropolis Hastings with Dynamic Coloring}

After completing a round of inference on all blue and red regions, we select a random offset uniformly from $[0, l]$ and shift all region boundaries by this offset. We then run another round of inference moves, and repeat the procedure shifting the region boundaries after each epoch. This solves two major problems we have in Chromatic Metropolis Hastings with a static coloring---inferring near-boundary events and applying joint moves to pairs of events separated by region boundaries.

In Figure~\ref{fig:actual_coloring}, let $A', B'$ be the current event hypothesis and $A, B$ the true events. With constant boundaries as in the graph, events $A',B'$ are in different regions and cannot be jointly moved to the other high probability true state $A, B$. However, with dynamic coloring, the region boundaries change based on the random offset. Whenever $A',B'$ are in the same region, the joint move can move them to the true state.

\section{Experimental Results} 
We evaluate inference methods for our seismic model on synthetic data generated by sampling from the model's distribution over possible worlds. We compute the accuracy of an inferred event hypothesis by comparing to the sampled ground truth. A bipartite matching is built between inferred and true events. We build the matching by looping through all true events and finding the closest inferred event to it. We add an edge between them if they are less than $12$ units apart in both time and distance. After an edge is added, the events on both sides of the edge are removed from the matching process, so that the degree of any event is at most $1$. After the matching is built, we report three key quantities: location error (average distance between matched events), precision (percentage of inferred events that are matched) and recall (percentage of true events that are matched). It is desirable to have a method with low error, high precision and high recall, which means the inference is more accurate in terms of both location and number of events. 

The setting of the world in our experiment is $x_{max} = 100, T = 240$. There are $n=4$ regions in our experiment. $\tau_{max} < \frac{T}{n}$, thus allows bi-coloring. For Chromatic methods, we run $500$ steps before switching to the other color. Naive parallel Metropolis Hastings algorithm can achieve a maximum $4$ times speed-up. Chromatic Metropolis Hastings algorithm can achieve a maximum $2$ times speed-up. The high-level distributed computing is built on PySpark. All the experiments are run on a quad-core computer.

\begin{figure*}[t!]
\centering
\captionsetup[subfigure]{labelformat=empty}
\begin{subfigure}{.245\textwidth}
  \centering
  \includegraphics[width=1.05\linewidth]{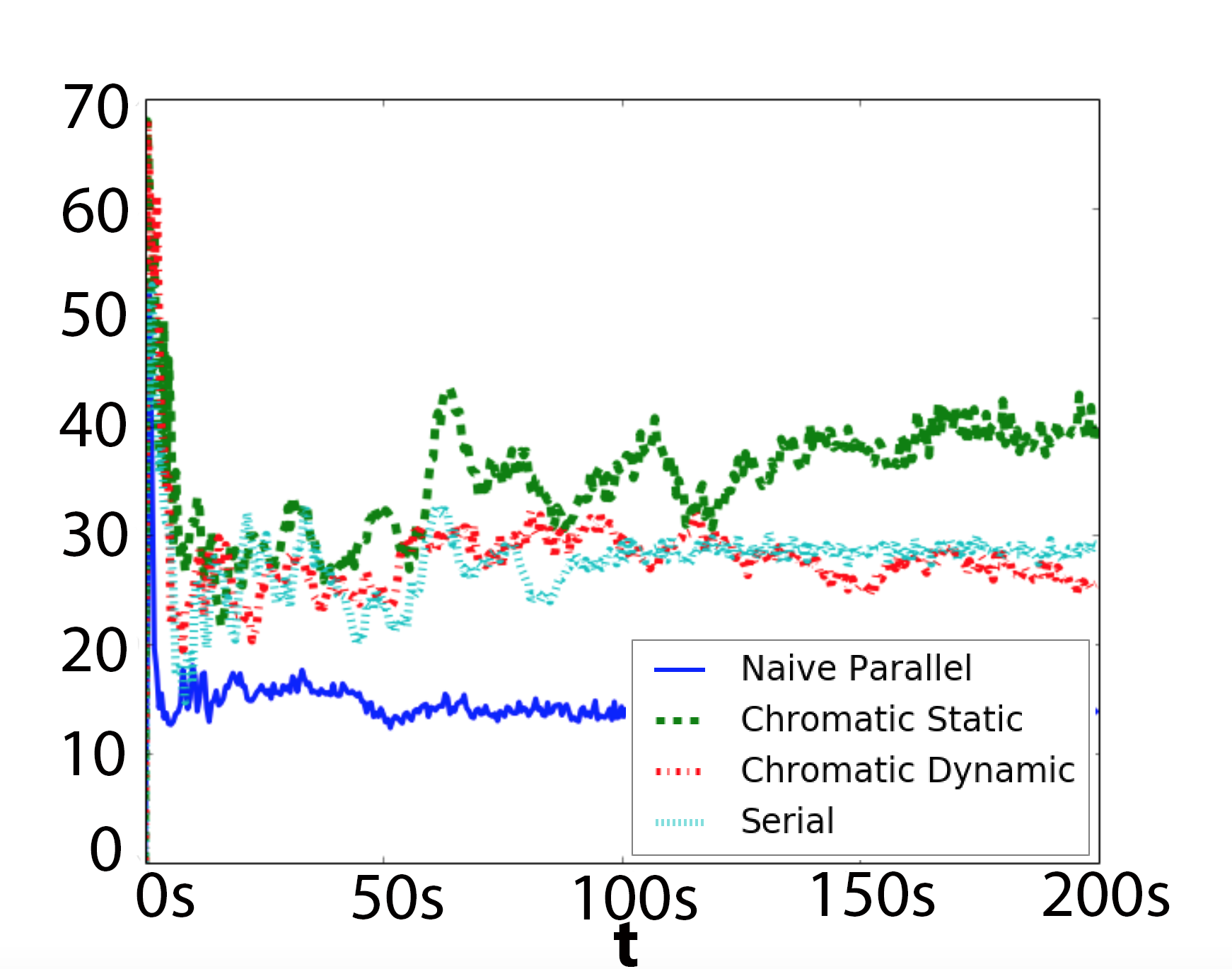}
  \caption{Location Error vs Time, lower is better}
\end{subfigure}%
\begin{subfigure}{.245\textwidth}
  \centering
  \includegraphics[width=1.02\linewidth]{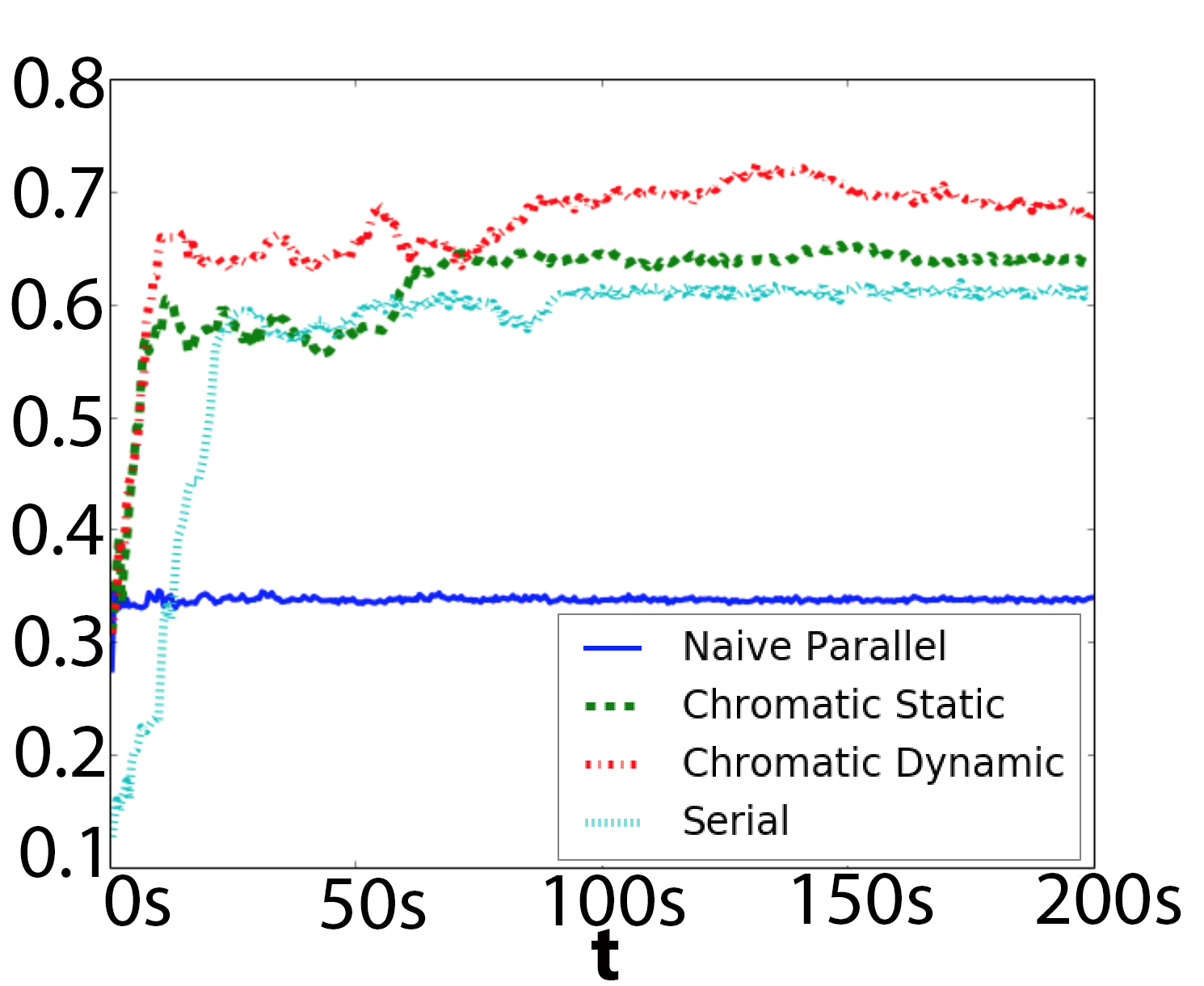}
  \caption{Precision vs Time, higher is better}
\end{subfigure}
\begin{subfigure}{.245\textwidth}
  \centering
  \includegraphics[width=1.015\linewidth]{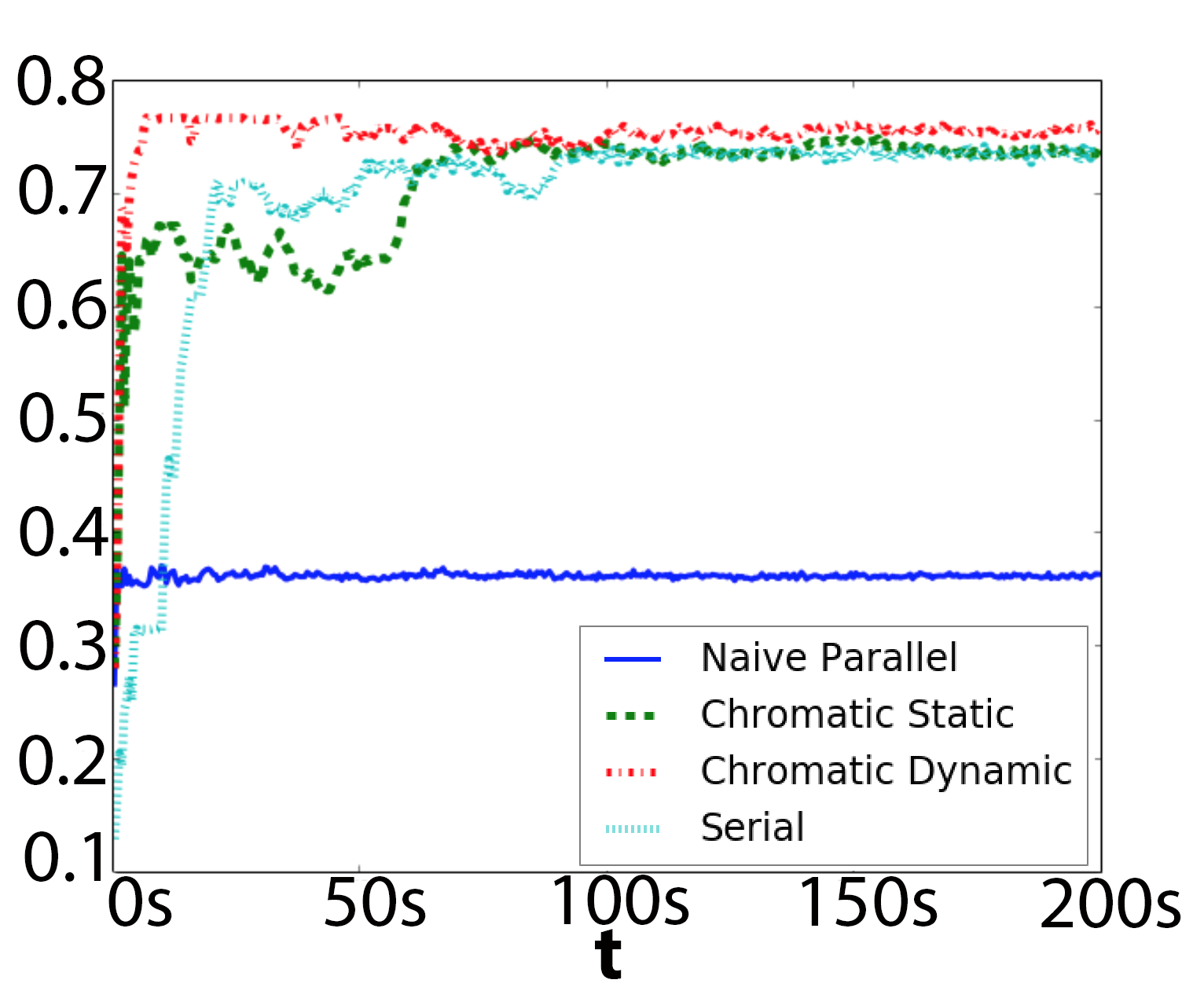}
  \caption{Recall vs Time, higher is better}
\end{subfigure}%
\begin{subfigure}{.245\textwidth}
  \centering
  \includegraphics[width=1.015\linewidth]{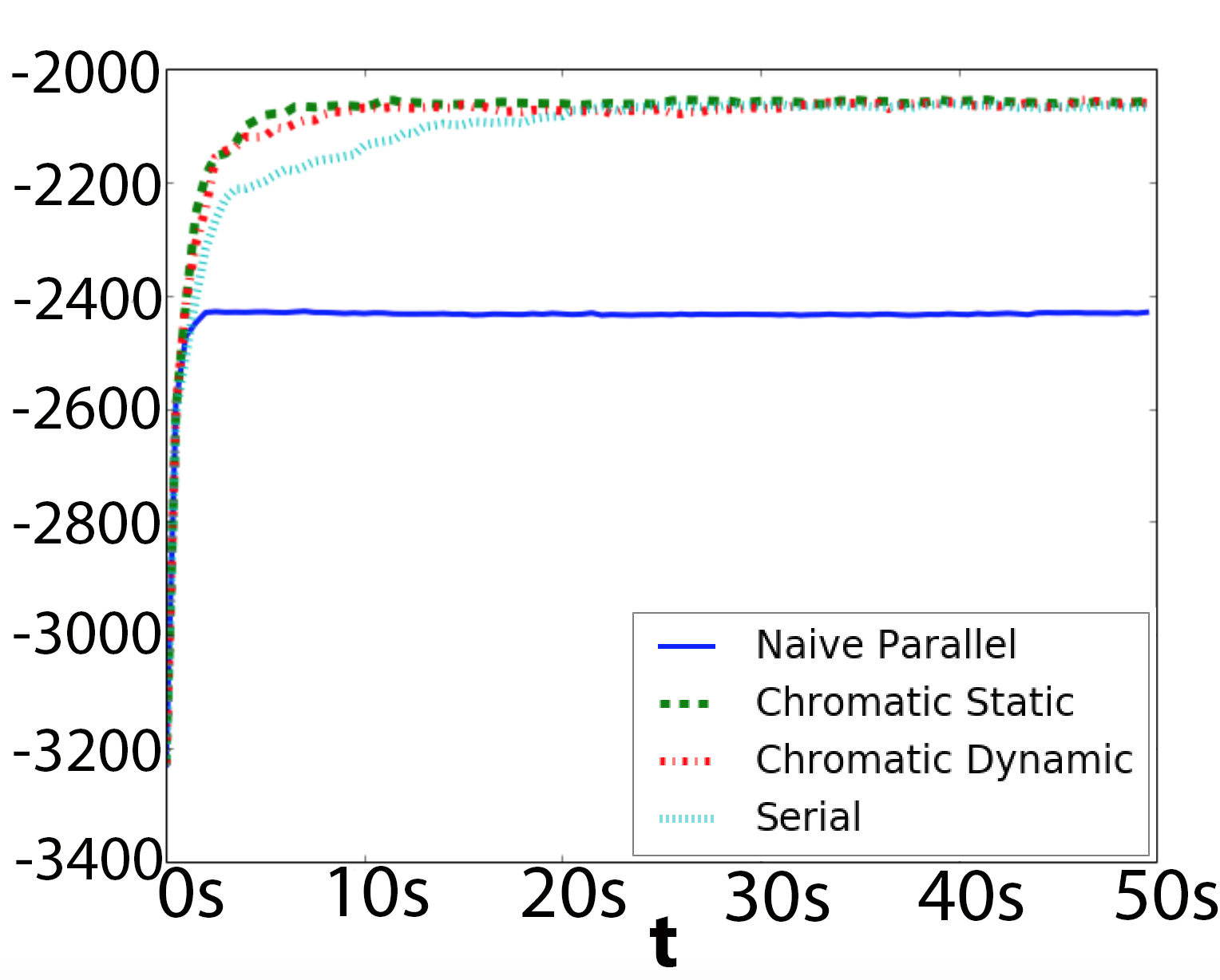}
  \caption{Log Probability vs Time, higher is better}
\end{subfigure}
\caption{}
\label{fig:metric}
\end{figure*}

Figure~\ref{fig:metric} shows location error, precision, recall, log probability versus time metrics for each algorithm. These metrics are generated based on the average result from $5$ random worlds. From the metrics, we see that Chromatic Dynamic, Chromatic Static and Serial are achieving similar overall final performance but Serial converges slower than the other two methods. Naive Parallel converges fastest but it has the lowest precision and recall. The low error of Naive Parallel is also an artifact of its low recall (generates more events than actual). In the metrics, Chromatic Dynamic has a slightly better precision and recall than Serial algorithm because of random variation. This motivates us to do confidence interval analysis of the three methods on more random worlds. 

\begin{figure*}
\centering
\begin{subfigure}{.31\textwidth}
  \centering
  \includegraphics[width=0.80\linewidth]{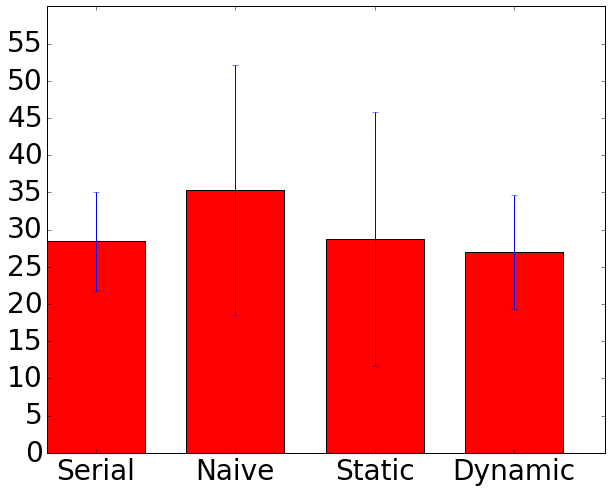}
  \caption{Location Error}
\end{subfigure}%
\begin{subfigure}{.31\textwidth}
  \centering
  \includegraphics[width=0.80\linewidth]{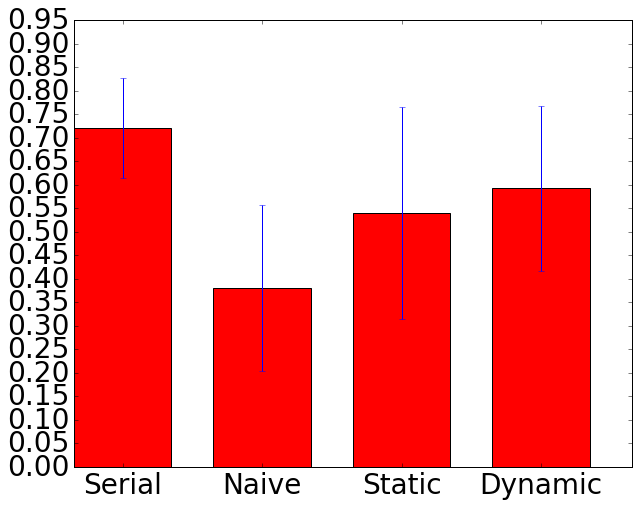}
  \caption{Precision}
\end{subfigure}
\begin{subfigure}{.31\textwidth}
  \centering
  \includegraphics[width=0.80\linewidth]{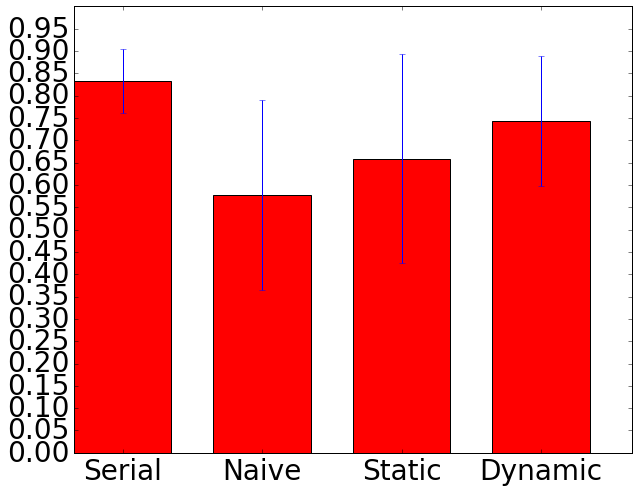}
  \caption{Recall}
\end{subfigure}%
\caption{Bootstrap Confidence Interval of Location Error, Precision and Recall}
\label{fig:bootstrap}
\end{figure*}

Using $50$ possible worlds sampled from the model, we run each algorithm $5$ times and measure the average precision, recall and error of our inference via the assumed ground truth. From metrics above, we can see that at $200$ seconds, all algorithms have safely converged. So in this experiment, we run longer than $200$ seconds for each algorithm to show the final performance after convergence. After we get $50$ sets of results, we calculate the mean location error, mean precision and mean recall for each algorithm and generate error bars around the mean using $95 \%$ bootstrap confidence intervals \citep{Efron1993}. See Figure~\ref{fig:bootstrap}. From the bar graph, we see that dynamic partitioning yields precision and recall statistically indistinguishable from serial MCMC, and significantly better than na\"ive parallelization.  

\section{Related Work}
When partitioning spatial models, the general idea is to group varaiables in a way that limits dependencies cross partitions. Previous approaches to exploiting model structure model to distribute learning \citep{Fugue,NIPS2014_5598} attempt to achieve data parallelism, model parallelism or both. However, this set of approaches is restricted to models expressible in a matrix form. 
Some previous parallel MCMC algorithms have been designed for specific models, such as mixture models \citep{d92223573bc5449a8e62b82194483d88} and topic models \citep{Newman:2009:DAT:1577069.1755845,Smola:2010:APT:1920841.1920931}. However, this set of algorithms is not applicable to other graphic models, such as the seismic model in our setting. Another parallel MCMC algorithm \citep{journals/corr/NeiswangerWX13} is more general, but it requires observation to be i.i.d. independent data points. Since we exploit a different kind of structure -- spatial/temporal locality in the observations themselves, our approach could be complementary to \citep{journals/corr/NeiswangerWX13} or parallel MCMC with other conditional independence assumptions in observations.

\section{Conclusion and Future Work}
We have introduced a novel approach for parallel MCMC inference exploiting value-dependent conditional independence induced by spatial structure rather than a fixed graphical model. Evaluating on a simple model of seismic events and signals, we find that the Chromatic Metropolis Hastings using Dynamic Coloring can achieve $\frac{n}{k}$ times speedup while still maintaining a similar level of error, precision and recall as Serial Metropolis Hastings, where $n$ is the number of spatial partitions and $k$ the number of colors. 

Future work involves extending this approach to production-scale models such as that of \citet{NIPS2010_4100}, and to other applications involving spatial object detection and/or localization. Also of interest would be methods for automatically detecting spatial independence relationships of the form we exploit, given formal descriptions of model structure, and selecting an appropriate partition of the latent space to maximize inference efficiency.

\bibliography{reference}
\bibliographystyle{aaai}

\end{document}